\documentclass{article}

\usepackage[final]{neurips_2021}
\usepackage{graphicx}
\usepackage{subfig}
\usepackage{amsmath}
\usepackage{multirow}
\usepackage{multicol}
\usepackage{hyperref}
\usepackage{makecell}
\usepackage[utf8]{inputenc}
\usepackage{algorithm}
\usepackage[noend]{algpseudocode}
\usepackage{siunitx}
\usepackage{booktabs}
\usepackage{svg}

\usepackage[utf8]{inputenc} 
\usepackage[T1]{fontenc}    
\usepackage{hyperref}       
\usepackage{url}            
\usepackage{booktabs}       
\usepackage{amsfonts}       
\usepackage{nicefrac}       
\usepackage{microtype}      
\usepackage{xcolor}         

\title{Curriculum Meta-Learning for Few-shot Classification}

\author{
  Emmanouil Stergiadis \\
  Booking.com\\
  \texttt{emmanouil.stergiadis@booking.com} \\
  \And
  Priyanka Agrawal \\
  Booking.com\\
  \texttt{pagrawal.ml@gmail.com} \\
  \And
  Oliver Squire \\
  Booking.com\\
  \texttt{oliver.squire@booking.com} \\
}

\begin{document}

\maketitle

\begin{abstract}
 We propose an adaptation of the curriculum training framework, applicable to state-of-the-art meta learning techniques for few-shot classification. Curriculum-based training popularly attempts to mimic human learning by progressively increasing the training complexity to enable incremental concept learning. As the meta-learner's goal is learning how to learn from as few samples as possible, the exact number of those samples (i.e. the size of the support set) arises as a natural proxy of a given task's difficulty. We define a simple yet novel curriculum schedule that begins with a larger support size and progressively reduces it throughout training to eventually match the desired shot-size of the test setup. This proposed method boosts the learning efficiency as well as the generalization capability. Our experiments with the MAML algorithm on two few-shot image classification tasks show significant gains with the curriculum training framework. Ablation studies corroborate the independence of our proposed method from the model architecture as well as the meta-learning hyperparameters.
\end{abstract}

\section{Introduction}
\label{sec:intro}

Meta-learning works \cite{Thrun1998} introduced a prominent  set of techniques that aim to learn transferable knowledge from seen tasks and thus can generalize well with few training examples for a range of unseen tasks and environments. These works can be broadly categorized into metric-based methods \cite{Vinyals2016,SnellSZ17,Sung2018},  model-based methods \cite{Memory2016,munkhdalai17a,mishra2018a} and gradient-based methods \cite{Imagenet2017,NicholReptile2018}. Most of these works train a model with a set of past tasks drawn from the same distribution, and thus similar in nature, to the new tasks. However, training a model on extremely data-scarce tasks that mirror the unseen ones limits its ability to learn a good base learner, even when more annotated samples were in fact available for the tasks using in meta-training. Some works \cite{NicholReptile2018,Lee2019} alleviate this by using  a larger support size for meta-training tasks, but this approach does not work consistently for all techniques (see Section \ref{sec:adaptive}) due to the discrepancy it introduces between meta-training and meta-testing. 
This presents a trade-off between  training  a better base  learner  versus  generalization  capabilities.

In this work, we propose a curriculum learning framework with a dynamic training schedule. This allows one to train a better base learner using a larger support set during initial epochs as well as adhere to the shot-size constraints towards the later epochs. While the use of curriculum learning for training machine learning models \cite{Bengio2009CurriculumL} is not new, the traditional curriculum learning methods use individual sample complexity and sort the samples respectively. 
To the best of our knowledge, this work is the first to propose a curriculum design for episodes/tasks in meta-learning approaches that uses support-size as a proxy for the difficulty of a meta-task. We conduct detailed experimental analysis (see Sections \ref{sec:setup} - \ref{sec:res}) with two few-shot image classification datasets to demostrate the efficacy of our proposed curriculum learning method. 

\section{Related Work}
\textbf{Meta-learning} works for few-shot classification problems aim to model the knowledge of known tasks in a form that is easily transferable to new, unseen classes with very small set of examples. These works can be broadly organized into three main categories: \\(i) metric-based  or `learning to compare' methods which rely on learning a model that can determine similarity between two examples. This can be used to match an unseen instance with the labeled instances \cite{Koch2015SiameseNN,Vinyals2016,SnellSZ17,Sung2018}; \\(ii) model-based or `good model initialization' methods attempt to learn a model that can be fine-tuned for novel classes with a limited number of examples and few parameter updates \cite{Memory2016,munkhdalai17a,mishra2018a}; \\(iii) gradient-based  or `learning an optimizer' methods aim at adjusting that optimization algorithm so that model can adapt with few gradient updates  \cite{finn17a-pmlr-v70,Imagenet2017,NicholReptile2018,antoniou2018how}. 

The last two categories can also be termed as initialization-based or 'learning to fine-tune' methods. While the initialization based methods aim to enable rapid adaptation with a limited number of training
examples for novel classes, these methods are shown \cite{chen2018a} to have limited  generalization owing to the shifts between base and novel meta-testing classes as well as an inefficient base learner initialization owing to the difficulty of meta-training tasks. This motivates our attempt towards improving the base-learner while eliminating any discrepancy between the meta-training and meta-testing setups. 
\\
\\
\textbf{Curriculum learning} \cite{Bengio2009CurriculumL} has been actively used for several tasks \cite{Hacohen2019OnTP} to allow for ordering of training examples based on their difficulty, thus facilitating incremental learning of complex concepts. Its applications span across problems ranging from single static tasks \cite{spitkovsky2010-baby} to continual learning \cite{Wu2021} and transfer learning \cite{Weinshall2018,2020Zhang}.\\
Recent work \cite{Wu2021} uses curriculum as well as meta-learning for continual relation extraction tasks to deal with catastrophic forgetting and order-sensitivity. However, unlike our method, the curriculum is limited to the traditional ordering of samples based on their complexity. 
To the best of our knowledge, our work is the first to propose a curriculum design for episodes/tasks in meta-learning approaches.
 It is important to note that while we proposed a curriculum learning schedule for gradient-based methods, our method can also be adapted for nearest-neighbor and metric-based methods.

\section{Proposed Approach}
\label{sec:method}
Similar to the gradient-based method of model-agnostic meta-learning (MAML) \cite{finn17a-pmlr-v70}, our work employs a meta-training phase that explicitly optimises a weight initialization $\theta$ on a set of tasks, such that the base learner initialized with those weights is able to adapt to any unseen task when provided with a few samples (typically 1 to 20) of each class. 

Unlike several meta learning works including MAML, we draw a clear distinction between the number of training examples at meta-test time (hereafter called $shot$) and the support size during meta-training (hereafter denoted by $K$). This distinction has also been made in recent works such as \cite{NicholReptile2018} that proposed meta-training with support size larger than the shot size i.e. $K > shot$. As the support set enables the meta-learner to adapt to a new task, we use the support size $K$ as a proxy for a task's difficulty. 

We propose that the support size $K$ should therefore be adapted throughout meta-training. Concretely, we opt for a larger support size in the initial stages of meta-training that will be progressively decreased to eventually match that of the test distribution ($shot$) towards the end. This is equivalent to a curriculum design for the training episodes/tasks, where the initial meta-training tasks are easier due to the availability of larger support set and as the meta-training progresses, the tasks are harder due to limited support set.  While several schedules can be designed to fit such as a general definition, we limit our experiments to a rather simple yet novel schedule:\\
At each stage, the support set size will be equal a multiplier  of the desired $shot$, where the multiplier value decreases with each stage. We first set the maximum multiplier value to a small natural number (which also corresponds to the number of curriculum stages) $M$ as a hyper-parameter. We then reduce this multiplier by 1 in equally spaced iteration intervals until it reaches the value of 1, at which point $K = shot$ and the meta-training setup exactly matches that of meta-test. This computation corresponds to the line 7 of Algorithm \ref{algo}. We denote the support size of each stage $i$ as $K_i$ where

\begin{equation}
    K_i= (M - i) * shot \quad \forall i \in[0, M)
\label{eq:k}
\end{equation}

Following MAML we do not employ early stopping; instead a predetermined number of steps \textbf{N} is set as a hyper-parameter. However the checkpoint that achieved the lowest validation error is used during meta-testing to address potential overfitting. We designate half the training steps to the last stage $K_M$ irrespective of the actual $M$ setting in order to stabilise the meta learner and eliminate any discrepancy between meta-training and meta-testing. Therefore each stage will train for \textbf{n\textsubscript{i}} steps where:

\begin{equation}
    n_i= 
\begin{cases}
    \frac{N}{2M - 2},& \text{if } i < M\\\\
    \frac{N}{2},              & \text{if } i = M
\end{cases}
\label{eq:n_m}
\end{equation}

\begin{algorithm}[H]
\caption{\label{algo}Curriculum MAML}
\begin{algorithmic}[1]
    \Require \textbf{$\mathcal{T}$}: distribution over tasks
    \Require $\alpha_0$, $\beta$: learning rate for inner and outer loop
    \Require  \textbf{N}: number of iterations, \textbf{M}: multiplier
    
\State randomly initialize $\theta$
\State \textbf{set} $K \leftarrow shot * M$
\State \textbf{set} $\alpha \leftarrow  \alpha_0 * \sqrt{M}$
\State \textbf{set} $n \leftarrow (N / 2) /  (M - 1)$
\For{$i$ \textbf{in} $range(1, N)$}
    \If{$modulo(i, n) = 0$}
        \State \textbf{update} $K \leftarrow max(K - shot, shot)$
        \State \textbf{update} $\alpha \leftarrow max( \sqrt{\alpha^2 - \alpha_0^2}), \alpha_0)$
    \EndIf
    \State Sample batch of tasks $\mathcal{T}_i$ $\sim$ $p(\mathcal{T})$
    \For{all $T_i$}
        \State Sample $K$ datapoints $D = \{x^j, y^j\}$ from $T_i$
        \State Evaluate the cross entropy loss $\nabla{_\theta}\mathcal{L}_{\mathcal{T}_i}(f_\theta)$ on $D$
        \State Compute adapted parameters with gradient descent $\theta_i \leftarrow \theta - \alpha\nabla{_\theta}\mathcal{L}_{\mathcal{T}_i}(f_\theta)$
        \State Sample $L$ datapoints $D' = \{x^j, y^j\}$ from $T_i$ for the meta-update
        \EndFor
    \State \textbf{end for}
    \State Compute adapted parameters $\theta \leftarrow \theta - \beta\nabla{_\theta}\sum_{\mathcal{T}_i \sim p(\mathcal{T})}\mathcal{L}_{\mathcal{T}_i}(f_{\theta_i})$ on $D'$
    \EndFor
    \State \textbf{end for}
\end{algorithmic}
\end{algorithm}

We provide the pseudocode for our method in Algorithm 1 which closely follows Algorithm 2 in \cite{finn17a-pmlr-v70}. Observe that the term $\nabla{_\theta}\mathcal{L}_{\mathcal{T}_i}(f_\theta)$ in the inner loop is estimated with a monotonically decreasing number of examples $K$ and thus possibly increasing variance as the curriculum progresses through its stages. 

In order to capitalize on the  reduced variance during the initial stages of curriculum training, we propose an annealing schedule where the learning rate is scaled proportionally to the square root of the support set's size as per equation:

\begin{equation}
    \alpha_i = \sqrt{M - i} * \alpha_0 \quad \forall i \in[0, M)
\label{eq:lr}
\end{equation}
This annealing corresponds to line 8 in Algorithm \ref{algo}. Similarly to \cite{2020t5}, we select this annealing schedule because it does not require information on the total number of steps a-priori unlike other popular schedules like triangular learning rate (\cite{Howard2018}). We do not employ warm up.

\section{Experimental Setup}
\label{sec:setup}
\textbf{Datasets:} We conduct experiments on two few-shot image classification tasks, Omniglot \cite{Lake1332} and miniImagenet \cite{Imagenet2017}. Omniglot consists of 1623 characters drawn from 50 different alphabets, with 20 samples provided for each class. 1100 of those character classes are used for training, 100 for validation and the remaining 423 for testing. Similar to \cite{Memory2016} we augment omniglot with rotations by multiples of 90 degrees. MiniImagenet consists of 100 classes, of which 64 are used for training, 12 for validation and 24 for testing purposes. No data augmentation is used. We release the code to reproduce our results at \textit{https://github.com/steremma/curr-meta-learning}.
\\ 
\\
\textbf{Base models:} We follow the setup from MAML\footnote{Our initial experiments with REPTILE also result in gains similar to shown in Section \ref{sec:res1} thus indicating applicability of proposed curriculum frameworks irrespective of the baseline meta-learner model.} \cite{finn17a-pmlr-v70}. Our base learners exactly follow MAML, both in the convolutional and the feed forward variant:

\begin{itemize}
    \item Our \textbf{convolutional Network} follows \cite{Vinyals2016} which employs 4 layers, each using 3x3 kernels and 64 filters on Omniglot or 32 filters on miniImagenet \cite{Imagenet2017}. Convolution is followed by batch normalisation and the ReLU activation function. Lastly we address overfitting using downsampling: a 2x2 max-pooling operation on miniImagenet and strided convolution on Omniglot.
    \\
    \item Our \textbf{Feed Forward Network} comprises of 4 hidden layers with sizes 256, 128, 64, 64 each including batch normalization and ReLU activation, followed by a linear layer.
\end{itemize}

While first order approximation of the Hessian matrix utilised in MAML's outer loop was shown to speed-up computation with minor performance degradation, we do not apply it on our experiments. All training schemes run for a total of $N = 60,000$ steps. These steps are allocated per curriculum stage as per Eq. \ref{eq:n_m}. The curriculum framework is applied over the implementation available in  \textit{learn2learn} repo\footnote{https://github.com/learnables/learn2learn}.
\\
\\
\textbf{Experiments:} We aim to investigate the following research questions related to our proposed approach:\\
\\
\textbf{RQ1:}
\textit{Does curriculum based training lead to an improved model for unseen tasks in comparison to baseline static meta-learner?}

We experiment with a range of curriculum stages $M = {20, 10, 5, 1}$. The setting with $M = 1$ corresponds to our baseline MAML training where there is no curriculum stage and the training support set size is same to that of the test.
Therefore, we treat $M=1$ as a baseline.
\\
\\
\textbf{RQ2:}
\textit{Do hyperparameters like learning rate, query set size etc. need to be adapted for each curriculum stage?}

We observe that the support set serves as the batch for the base learner and therefore enlarging the support set should ideally stabilize the gradients. This could enable the use of larger learning rate to accelerate training. We explore this hypothesis by experimenting with adaptive learning rate per equation \ref{eq:lr}.

Furthermore, it is not obvious whether the query set should always match the support set in terms of size in each stage, or instead stay statically equal to its test-time configuration. To answer this, we experiment with both adaptive query set size $L = K$ and the static one $L = shot$. 
\\
\\
\textbf{RQ3:}
\textit{Is the effect of curriculum meta learning limited to specific training configurations?}

We investigate whether improved performance is limited to specific configurations such as:
\begin{itemize}
    \item The base learner.
    \item The dataset's complexity.
    \item The desired $shot$ configuration.
\end{itemize}
To answer this, we attempt our schedule on both base learner setups used in the original paper \cite{finn17a-pmlr-v70}, namely a fully connected and a convolution network, and both datasets (miniImagenet \& Omniglot) with $shot=1$ or $shot=5$ examples.
\\
\\
\textbf{RQ4:}
\textit{Is gradual curriculum  important? }

Several existing works (\cite{NicholReptile2018}, \cite{Lee2019}) use a larger support set during meta-training in comparison to the $shot$ size used during meta-testing. We isolate the effect of curriculum by decoupling the effect of using bigger support size in our ablation study.


\section{Results and Discussion}
\label{sec:res}
\subsection{Impact of our proposed method:}
\label{sec:res1}

As highlighted in our research question \textbf{RQ1}, we compare the performance of proposed curriculum learning framework with that of the baseline MAML model. We start by identifying a reasonable setting for $M$ i.e. the number of curriculum stages. This analysis is done on the miniImagenet dataset with a 5-way 5-shot setup using a convolutional network as the base learner.

\begin{figure}[t]
\centering
\includegraphics[width=10cm, scale=1]{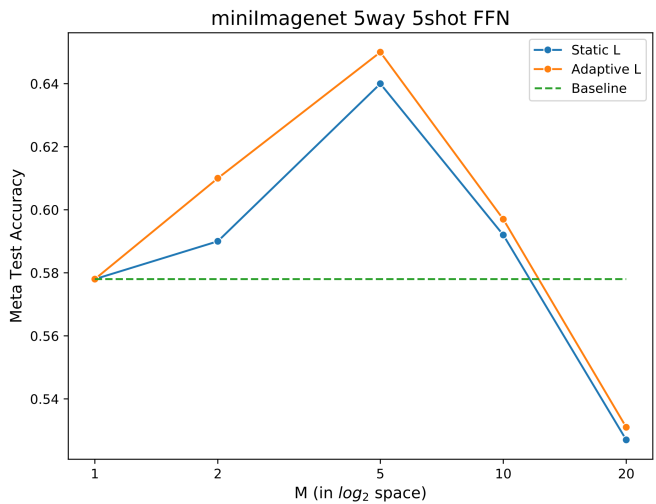}
\caption{\label{fig:optimalM}Meta Test performance on  5-way 5-shot miniImagenet using a Convolutional Network as the base learner. We attempt both a static ($L = shot$) and an adaptive ($L = K$) query size. For both configurations, the multiplier ($M$) of 5 is optimal, while adaptive $L$ slightly outperforms the static one consistently across the range of multiplier values.}
\end{figure}

As we can see in Fig. \ref{fig:optimalM}, curriculum-based training significantly outperforms the baseline setup for a large range of settings $M$. 
For this dataset, we find that a reasonable choice is $M=5$ and we continue to use this setting for our following experiments. Fig. \ref{fig:optimalM} additionally illustrates a comparison between static and adaptive query set size. More details on this can be seen in Section \ref{sec:adaptive}

Interestingly, even a smaller value of $M=2$ shows improved performance in comparison to the baseline\footnote{In the 5-shot setting our reproduction of baseline diverges from the one reported in \cite{finn17a-pmlr-v70}. However values of $M$ around 5 result in test accuracy even higher than the one reported in the original paper. Fortunately, our baseline experiments exactly match the numbers reported in \cite{finn17a-pmlr-v70} for all other experimental setups}. This is important in practice, because in several applications there is a practical limitation for the maximum $M$ one can use: The minimum amount of labeled samples required for each class participating in meta-training is increased from $shot$ to $M * shot$. This is for example the reason we do not report 5-shot performance on Omniglot, as this would require 25 samples per class in the first stage when only 20 are available\footnote{Ideally, one can conduct curriculum training for 5-shot setting on Omniglot by either allowing for  oversampling or reducing the number of curriculum stages}. In many applications this limitation does not pose an issue. For example, consider a setup in which candidate labels are dynamically added to a  classification system over time. In such cases, a practitioner might have plenty of labelled samples for historical classes at their expense but only \textbf{shot} samples for the newly added classes. In such cases the increased requirements for historical labelled data do not limit the selection of $M$.

On the other hand, it is important to note that for very high values of $M=\{10,20\}$ the performance starts dropping. 
This possibly happens because it becomes difficult for the model to adapt to the meta-testing environment due to its exposure to high amounts of training data during meta-training.

\begin{figure}[t]
\centering
    \subfloat[\centering Training Accuracy]{{\includegraphics[width=6.6cm, scale=1]{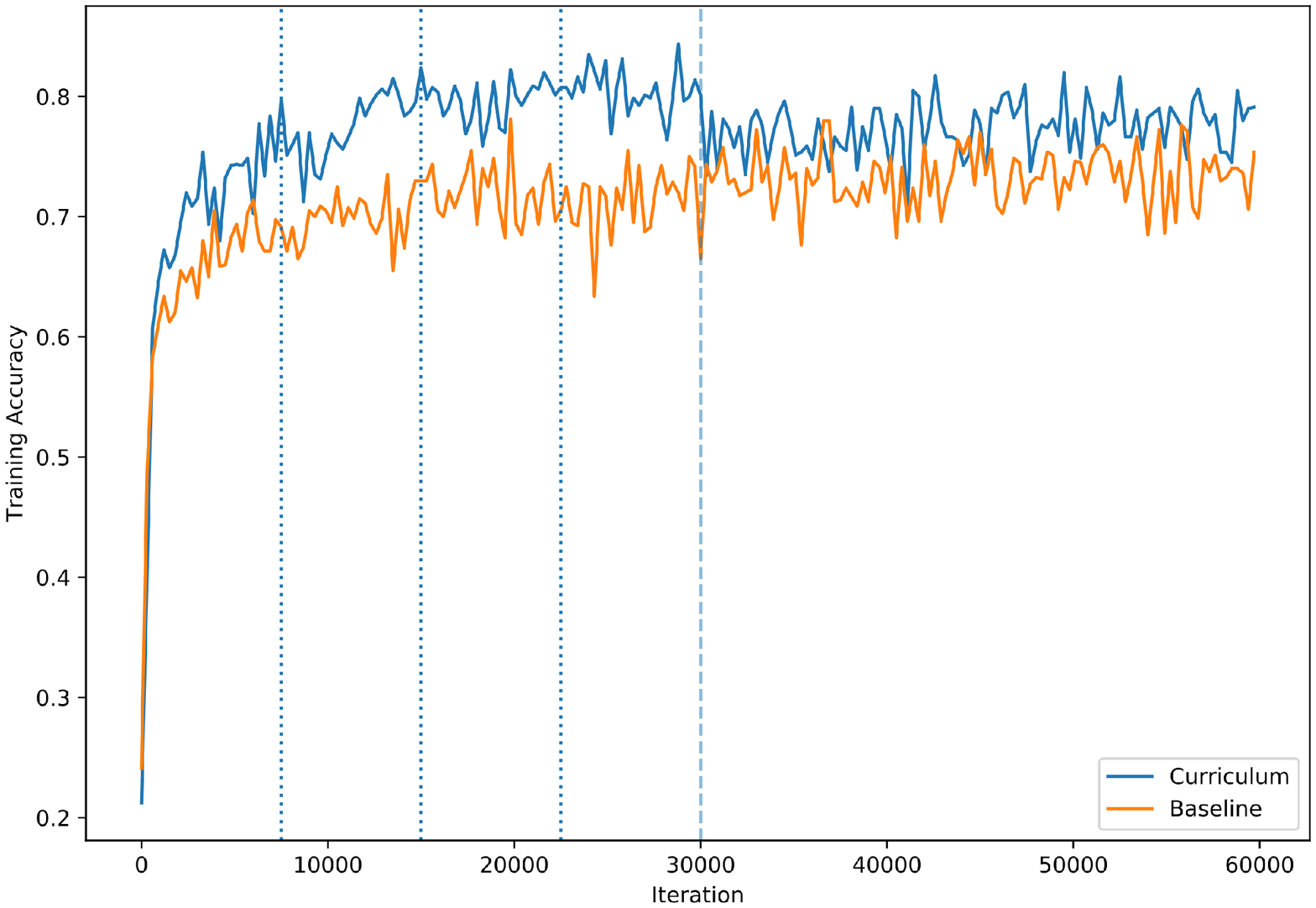}}}
    \qquad
    \subfloat[\centering Validation Accuracy]{{\includegraphics[width=6.6cm, scale=1]{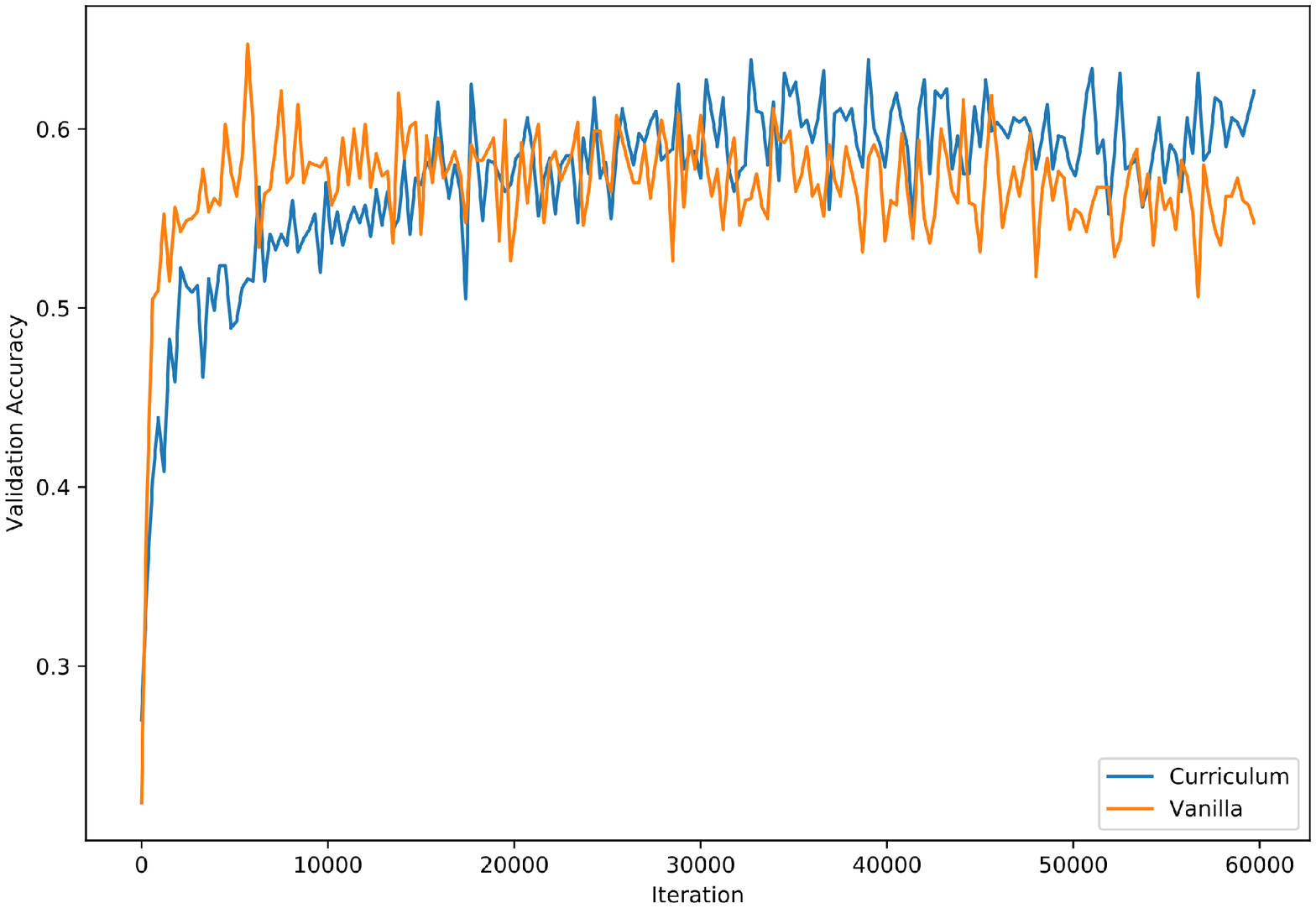}}}
\caption{\label{fig:train-accuracy}{Training and Validation accuracy  for miniImagenet 5-way 5-shot with $M= 5$}. Vertical lines denote stage switching. Prior to the last stage the base learner is trained on $K > shot$ examples which is reflected in the much higher training accuracy.}
\end{figure}

This trade-off between the base learner optimization and generalization capacity is illustrated by juxtaposing the training accuracy curves in Fig. \ref{fig:train-accuracy}. During the first curriculum stages, the support set is the largest, resulting in very high training accuracy, which does not translate to validation accuracy because of the huge discrepancy between training and validation. In contrast, when entering the last stage where $K=shot$ the training accuracy unsurprisingly takes a hit as the number of training examples halves, however it stays above the baseline that was training on this size all along. At the same time the validation accuracy increases as the gap between the training and validation setups has vanished. 

\subsection{Generality of the proposed method}
\label{sec:base}
As per our \textbf{RQ3}, we investigate whether the results seen in Section \ref{sec:res1} continue to hold with different base learners. For this purpose we design 2 setups on Omniglot, only differing in their base learner configuration:
\begin{itemize}
    \item OmniGlot 5-way 1-shot with a Feed Forward Network as the base learner. 
    \item OmniGlot 5-way 1-shot with a Convolutional Neural Network as the base learner. 
\end{itemize}

\begin{table*}[ht]
\setlength{\tabcolsep}{8pt}
\centering
\begin{tabular}{lcccc}
\multirow{3}{0pt}{Model} & \multicolumn{2}{c}{\textbf{Omniglot}} & \multicolumn{2}{c}{\textbf{miniImagenet}}\\
& {FFN} & {CNN} & {CNN} & {CNN}\\
& {5-way} & {5-way} & {5-way} & {5-way}\\
& {1-shot} & {1-shot} & {5-shot} & {1-shot}\\
\toprule
\textbf{Baseline MAML} & 0.90& 0.99& 0.58& 0.48\\
\textbf{Curriculum learning} & & & &\\
\quad $L = shot$ (static)&\makecell{0.92\\(+.02)}&\makecell{\textbf{1.00}\\(\textbf{+.01})} &\makecell{0.64\\(+.06)} &\makecell{0.51\\(+.03)}\\[12pt]
\quad $L = K$ (adaptive)&\makecell{\textbf{0.94}\\(\textbf{+.04})}&\makecell{\textbf{1.00}\\(\textbf{+.01})} & \makecell{\textbf{0.66}\\(\textbf{+.08})}& \makecell{\textbf{0.54}\\(\textbf{+.06})}\\[12pt]
\textbf{Static Support Size} & & & &\\
\quad $K = M * shot$, $L = shot$ &\makecell{0.87\\(-.03)}&\makecell{0.98\\(-.01)} & \makecell{0.51\\(-.07)}& \makecell{0.45\\(-.03)}\\[12pt]
\quad $K = M * shot$, $L = K$ &\makecell{0.91\\(+.01)}&\makecell{0.99\\(+.01)} & \makecell{0.55\\(-.03)}& \makecell{0.46\\(-.02)}\\[12pt]
\bottomrule
\end{tabular}
\bigskip
\caption{Results on the OmniGlot and miniImagenet datasets. All variants use a multiplier $M = 5$. Our static variants use $K = M * shot$ throughout training. For the miniImagenet results, we report baseline results as reproduced in our environment which in the 5-shot case diverge from the ones reported in \cite{finn17a-pmlr-v70}. The original paper mentions a 5-shot test accuracy of \textbf{0.63} instead of \textbf{0.58} as reported here. However, the deltas would likely lie in the same range as our variants are based on the same implementation as the reported baseline.}\label{tab:omniplusimage}
\end{table*}

The results are shown in the first three rows of Table \ref{tab:omniplusimage}. We first report the baseline version we reproduced which closely mirrors the results reported in the original MAML paper. The following two rows correspond to different curriculum variants, both of which outperform the baseline irrespective of the base learner. This result confirms that our method is largely agnostic to the choice of base learner.

Subsequently, we explore whether the observed performance boost is related to the complexity of the dataset, or the desired $shot$ setting. Also in Table \ref{tab:omniplusimage} we show results on miniImagenet using the convolutional base learner for $shot=1,5$. In both setups, we observe exactly the same behaviour as in Omniglot, i.e. the curriculum schedule results in improved performance.  These findings corroborate that our method is robust to the base learner, the complexity of the dataset and the desired $shot$ configuration.

\subsection{Performance with adaptation of hyperparameters:} 
\label{sec:adaptive} We analyze if setting hyperparameters in accordance to the support set's size $K$ at a given curriculum stage yields further improvement. For this study, we select two prominent hyperparameters: \textbf{learning rate} $\alpha$ and \textbf{query set size} $L$. 
\\
\\
\begin{figure}[ht]
\centering
\includegraphics[width=10cm, scale=1]{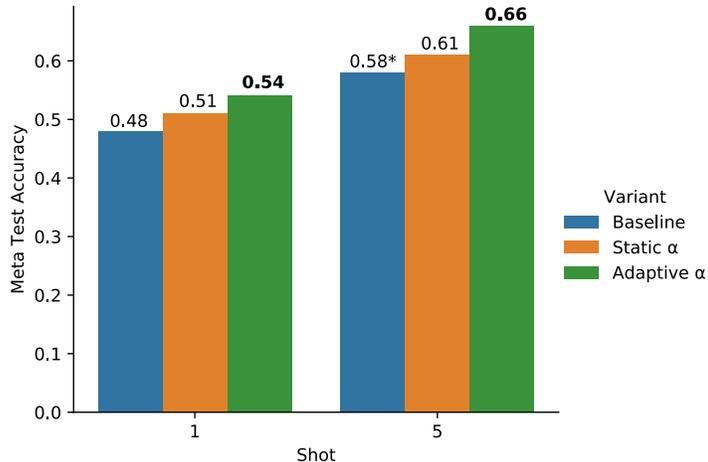}
\caption{\label{fig:lr}{Meta-Test performance on 5-way 5-shot miniImagenet using a Convolutional Network as the base learner. We explore the effect of annealing the learning rate proportionally to the square root of the support set's size per Equation (\ref{eq:lr}).}}
\end{figure}

\textbf{Adapting learning rate ($\alpha$):} As discussed in section \ref{sec:method}, the larger support set used in early stages of the curriculum should yield a less noisy gradient to the base learner in the inner loop. In addition, we hypothesize that increasing the learning rate of the base learner could lead to faster and potentially better optimization. Specifically we attempt to scale the learning rate as per equation \ref{eq:lr} using the base $\alpha_0 = 0.5$ as in the original implementation. While curriculum learning with static learning rate already performs better than the baseline,  we found that annealing provides further consistent gains in terms of test accuracy as shown in Fig. \ref{fig:lr}. Therefore, throughout all experiments illustrated in our paper we use an adaptive learning rate. 


\textbf{Adaptive Query set size ($L$): }
For each of the setups listed in Section \ref{sec:base}, we first report the baseline version we reproduced, which closely matches the results reported in the original MAML paper. We then report performance for two variations of our proposed curriculum schedule: keeping the query size $L = shot$ throughout training, or adapting it to $L = K$ for each stage of training. Comparing the rows under the \textbf{Curriculum learning} header in Table \ref{tab:omniplusimage} shows that adapting the query set size consistently increases gains across experimental setups.

\subsection{Effect of statically larger support size}
Lastly, for all setups, we report ablation numbers where we train without curriculum, but instead with a larger support size equal to $K = M * shot$ (the maximum value used in the curriculum variations) kept constant throughout training. This is to explore whether our improvement comes from the schedule, or simply by using a support size higher than shot as reported in \cite{NicholReptile2018,Lee2019}. There are again two variations on this ablation, where the query size is statically equal to $shot$, or to $K$ corresponding to the last two rows in Table \ref{tab:omniplusimage}. Our ablation reveals that neither version achieves the improved performance of curriculum meta-learning. The observed gains should therefore be attributed to the proposed schedule on the support set size and not its absolute size at a given stage.   

 Additionally, since statically using a higher K and/or L does not yield this improvement we conclude that it is indeed the adaption of those parameters and not their maximum size that results in the performance boost.

\section{Conclusions}
We present a simple, yet effective curriculum method to boost few-shot classification performance within the meta-learning framework. We apply the proposed method on top of MAML, a well known meta learning algorithm, and show consistent gains across base learners, datasets and $shot$ configurations. Furthermore, our method is conceptually applicable to other particular meta learning frameworks and can be easily tested in combination with modern, state-of-the-art meta learning algorithms. To facilitate future work we make our source code publicly available.



\bibliography{egbib}
\bibliographystyle{achemso}
\end{document}